\documentclass[letterpaper, 10 pt, conference]{ieeeconf}
\IEEEoverridecommandlockouts    
\usepackage[textwidth=4cm,disable]{todonotes}

\usepackage{subcaption}
\usepackage{tabularx,booktabs}

\usepackage[mode=buildnew]{standalone}
\usepackage{tikz}
\usepackage{pgfplots}
\pgfplotsset{compat=newest}
\usepgfplotslibrary{groupplots}
\usetikzlibrary{patterns}
\usepackage{amsmath}
\usepackage{siunitx}
\usepackage{hyperref}
\usepackage{flushend}

\newcommand*\rd{d^{AB}}

\usepackage{graphics}           
\usepackage{times}              
\usepackage{amsmath}            
\usepackage{amssymb}            
\usepackage{graphicx}
\usepackage{algorithm}
\usepackage[noend]{algpseudocode}
\usepackage{booktabs}
\usepackage{color}

\definecolor{instructioncolor}{rgb}{.5,.5,.5}

\usepackage[font=small]{caption}

\def\secref#1{Sec.~\ref{#1}}
\def\figref#1{Fig.~\ref{#1}}
\def\tabref#1{Tab.~\ref{#1}}
\def\eqref#1{Eq.~(\ref{#1})}

\makeatletter
\usepackage{xspace}
\DeclareRobustCommand\onedot{\futurelet\@let@token\@onedot}
\def\@onedot{\ifx\@let@token.\else.\null\fi\xspace}
 
\def\ie{i.e\onedot}

\def\etal{{et al}\onedot}
\def\etalcite#1{\etal~\cite{#1}}

\makeatother

\usepackage{array}
\newcolumntype{L}[1]{>{\raggedright\let\newline\\\arraybackslash\hspace{0pt}}m{#1}}
\newcolumntype{C}[1]{>{\centering\let\newline\\\arraybackslash\hspace{0pt}}m{#1}}
\newcolumntype{R}[1]{>{\raggedleft\let\newline\\\arraybackslash\hspace{0pt}}m{#1}}

\renewcommand{\b}[1]{\mbox{\boldmath$#1$}}

\newcommand{\m}[1]{{\mbox{{\sffamily\slshape{#1\/}}}}}

\newcommand{\bu}{\b u}

\newcommand{\bx}{\b x}

\newcommand{\bz}{\b z}

\title{\LARGE \bf Resource-Aware Collaborative Monte Carlo Localization \\
with Distribution Compression}

\author{Nicky Zimmerman  \and Alessandro Giusti  \and J\'er\^ome Guzzi
  \thanks{All authors are with the Dalle Molle Institute for Artificial Intelligence (IDSIA), USI-SUPSI.}%
} 

\begin{document}   
\maketitle
\thispagestyle{empty}
\pagestyle{empty}

\begin{abstract}
  %
  Global localization is essential in enabling robot autonomy, and collaborative localization is key for multi-robot systems.
  In this paper, we address the task of collaborative global localization under computational and communication constraints. We propose a method which reduces the amount of information exchanged and the computational cost. We also analyze, implement and open-source seminal approaches, which we believe to be a valuable contribution to the community.  
  We exploit techniques for distribution compression in near-linear time, with error guarantees. 
  We evaluate our approach and the implemented baselines on multiple challenging scenarios, simulated and real-world. Our approach can run online on an onboard computer. We release an open-source C++/ROS2 implementation of our approach, as well as the baselines.\footnote{https://github.com/idsia-robotics/Collaborative-Monte-Carlo-Localization}
\end{abstract}

\section{Introduction} \label{sec:intro}

Globally localizing in a given map is essential for enabling robot autonomy. Localization becomes extremely challenging when 
the environment is highly symmetric, featureless or very dynamic. Human-oriented indoor environments such as office buildings, contain high degree of geometric symmetry due to repetitive structures. In addition, the readily-available map representations such as floor plans are lacking in details, resulting in seemingly identical structures. Relying solely on geometric features may result in localization failure, leading researchers to exploit additional sources of information. 
RFID~\cite{joho2009icra} and Wi-Fi signal strength~\cite{ito2014icra} can be used to improve pose estimation, 
as well as textual cues~\cite{zimmerman2022iros, cui2021iros}. Another venue is utilizing semantic 
information~\cite{atanasov2015ijrr, hendrikx2021icra,joho2009icra}, harnessing the significant progress in the fields of semantic understanding.
However, single-robot localization can still fail, and recovering from a localization failure in the single robot scenario, mostly involves a human in the loop, in particular when navigating with erroneous pose estimation raises safety concerns. In contrast, in a multi-robot setup, a poorly-localized robot can recover if assisted by a well-localized robot. 

In the last decade, multi-robot systems have become more prevalent, introducing a new challenge of multi-robot localization. As opposed to single-robot localization, where a robot estimates its pose based on its own sensing, in a collaborative setting a robot can make use of information it receives from other agents. A recurring pattern in the related research is using the state estimation of one robot to improve
 the localization of another robot upon detection~\cite{barea2007itssa, fox2000ar,prorok2012iros}. This information exchange  
 involves broadcasting the belief of the robots, a distribution that is often approximated by a particle filter in the case of 
 global localization. When the area the robots operate in is large, the particle set representing the belief can easily reach tens of 
 thousands of particles. The naive approach of simply broadcasting all of the particles and 
 then integrating them into another robot's belief~\cite{prorok2012iros},
  is computationally costly and has high bandwidth consumption. Therefore, a compressed representation
   of the belief is imperative and has been proposed in the past~\cite{prorok2012iros}. In this paper, we analyse the computational complexity related to compressing, exchanging, and fusing robots' beliefs for collaborative localization.
   
 
\begin{figure}[t]
  \centering
  \includegraphics[clip, width=\linewidth]{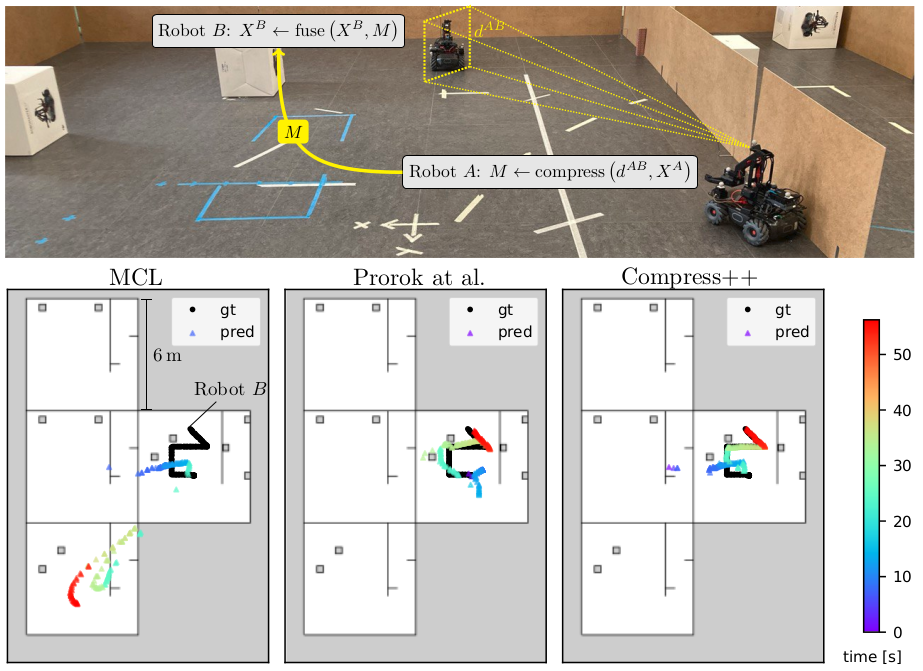}  
  \caption{Top: two robots during one experimental run when robot $A$ detects robot $B$. Bottom: localization of robot $B$ in the same run using 3 methods; prediction, color-coded for time, is plotted against ground truth position (black). Two failures, using non-collaborative MCL and Prorok ~\etalcite{prorok2012iros}, and a successful convergence after \SI{20}{s} with our collaborative localization approach (Compress++).}
  \label{fig:motivation}
\end{figure}

The main contribution of this paper is a novel approach to collaborative global 
localization~(\figref{fig:motivation}) which reduces the amount of data communicated and the computational cost. 
Additionally, we provide a unified overview and thorough analysis of alternative approaches to compress belief exchange. 
Furthermore, we release an open-source C++/ROS2 implementation for seminal works, as baselines for collaborative localization. 
While the methods are capable of operating in multi-robot scenarios, our experiments focus on the case of two collaborating robots, 
where we show that our approach 
 (i) improves collaborative localization, 
 (ii) decreases the required bandwidth, 
 (iii) reduces the computational load,
 (iv) runs online on an onboard computer. 

 The paper is organized as follows. \secref{sec:related} provides a summary of related research in collaborative localization.
 In \secref{sec:approach}, we present our novel approach for collaborative global localization with distribution compression, as well as
overview of alternative approaches. We analyze the complexity of distribution compression, communication and fusion in \secref{sec:CA}. 
We introduce our experimental setup in \secref{sec:setup}. In \secref{sec:exp}, we evaluate the performance of our approach against several
baselines, and offer insights regarding their different behaviors.  

\section{Related Work} \label{sec:related}  


 Map-based localization is an integral part of enabling the autonomy of mobile robots~\cite{cadena2016tro, thrun2005probrobbook}. 
 Global localization for a single robot is widely-researched, and probabilistic 
 methods~\cite{dellaert1999icra, fox1999jair, thrun2005probrobbook} have gained popularity due to their robustness. 
 The growing interest in multi-robot systems, presented a new challenge of collaborative multi-robot localization. 
 While the term multi-robot localization sometimes refers to relative positioning~\cite{howard2002iros, martinelli2005icra}, 
 we focus on cooperative global localization in a given map~\cite{fox2000ar}. 

 Multi-centralized approaches~\cite{nerurkar2010iros, nerurkar2009icra, roumeliotis2000tro}, where each robot maintains the belief for 
 all robots, generally do not scale well with the number of robots. In decentralize approaches, each robot estimates only its own state 
 and integrates relative observations from other robots when available. 

Similarly to the single robot case~\cite{atanasov2015ijrr, hendrikx2021icra,joho2009icra, zimmerman2023ral, zimmerman2023iros}, leveraging semantic scene understanding~\cite{cheng2020cvpr, he2017iccv-mr, sodano2023icra, bochkovskiy2020arxiv} was adopted for multi-robot localization. A common approach to collaborative localization relies on robot detection, where one robot can sense another robot. 
 Fox~\etalcite{fox2000ar} propose a factorial representation where each robot maintains its own belief, and the belief of different robots 
 are assumed to be independent from each other. When one robot detects another, the detection model is used to synchronize their beliefs. 
 However, no analysis is provided about the processing requirements and the experiments only cover one scenario. Barea~\etalcite{barea2007itssa},
 propose a system for collaborative localization based on Monte-Carlo localization~\cite{dellaert1999icra} framework, but do not include
 an explicit detection model.

 Wu~\etalcite{wu2009icrb} present an improvement to 
 Fox~\etalcite{fox2000ar}, where they consider whether the belief should be updated upon detection, by comparing the entropy of both robots' beliefs. They too, do not provide information about the computational and bandwidth requirement for the information exchange. 
 {\"O}zkucur~\etalcite{ozkucur2009robocup} introduce an approach to collaborative localization where robots can be detected but 
 not identified, but no details are given about the  complexity of the approach. Prorok~\etalcite{prorok2011iros} first detail a naive 
 approach of synchronizing the beliefs of two robots, each with N particles, with $O(N^2)$ complexity. 
 In their later work~\cite{prorok2012iros}, they provide a clustering algorithm to reduce the complexity to $O(NK)$, 
 where K is the user-defined number of clusters. In all works by Prorok~\etalcite{prorok2012iros, prorok2011iros}, the experimental setup does not include exteroceptive sensing and the utilization of a map, while we explore the contribution of belief exchange in the framework of a range-based MCL. With the exception of Prorok's works, the constrains imposed by limited compute and  bandwidth are largely unaddressed. Furthermore, the cited works present limited results for their individual approach, without comparing them against other baselines.  
 In our work, we provide thorough analysis for the computational and communication requirements of 
 seminal works~\cite{fox2000ar,prorok2012iros}, as well as present resource-aware alternative detection models. We benchmark the various approaches on several environments, and offer insights regarding their performance.


\section{Approach} \label{sec:approach}

We aim to globally localize a team of robots in a given map. In Sec.~\ref{sec:MCL}, we give a brief overview of 
Monte Carlo localization~\cite{dellaert1999icra}. We describe in \secref{sec:CMCL} a common approach to distributed multi-robot localization, 
where beliefs are exchanged when robots detect each other, including the concept of reciprocal sampling. In \secref{sec:DC}, we introduce our novel approach and, in \secref{sec:baselines}, we give an overview of alternative methods.

\subsection{Monte Carlo Localization}\label{sec:MCL}

Monte Carlo localization~\cite{dellaert1999icra} is a recursive Bayesian filter that estimates a robot's state $\b{x}_t$ at discrete time $t\in \mathbb{N}$, given map $m$, odometry readings $\b{u}_t$, and sensor readings $\b{z}_t$. We define the robot's state $\b{x} = (x, y, \theta)^\top$  as an horizontal position $(x, y) \in  \mathbb{R}^2$ and orientation $\theta \in [0, 2 \pi)$; the map $m$ is an occupancy grid map~\cite{moravec1989sdsr}; and the sensor measurement  $\b{z}\in   \mathbb{R}_{\ge0}^K$ has $K$ beams.
A particle filter represents the robot's belief $p(\bx_t | \bu_{1:t}, \bz_{1:t}, m)$ as a set $X_t = \{s_{t, i}, i=1, \dots, N\}$
where each particle \mbox{$s_{t, i}=\left(\bx_{t, i}, w_{t, i}\right)$} assigns weight $w_{t, i}$ to a sample state $\bx_{t, i}$. 

The proposal distribution $p(\bx_t | \bx_{t-1}, \bu_t)$ is sampled when odometry $\bu_t$ is available, using a motion model that considers the robot' kinematics and odometry noise $\sigma_{\mathrm{odom}}$. For each observation, the particle filter is updated based on the sensor model, where weight is assigned to each particle according to the likelihood of the observation given 
its state, \ie, $w_{t, i} = p(\bz_t | \bx_{t, i}, m)$. As observation model $p(\bz | \bx, m)$, we use a beam-end model~\cite{thrun2005probrobbook} with standard deviation $\sigma_{\mathrm{obs}}$. 
We perform the update only when the robot moves at least for a trigger distance $(\delta_{xy}, \delta_{\theta})$ to avoid repeated integration of the same observation.
In the resampling step, particles are selected based on their assigned weight: we resample a particle set of size $N$ using low-variance resampling~\cite{thrun2005probrobbook} 
with an efficiency coefficient $N/2$. 

\subsection{Collaborative Monte Carlo Localization}\label{sec:CMCL} 
Collaborative Monte Carlo localization extends the single robot MCL (\secref{sec:MCL}) by incorporating information from other robots into a robot's belief. Each robot locally runs an independent MCL that integrates its own sensor and odometry readings.
When robot $A$ detects robot $B$ at time $t$, it estimates the \emph{relative} position $\rd_t \in \mathbb{R}^2$  of $B$ and then sends message $M$ to $B$ with information about its state $X_t^A$ and detection $\rd_t$,  as illustrated in \figref{fig:motivation}.
$B$, upon receiving the message, updates the weights of its particles~\cite{fox2000ar}, fusing information:
\begin{equation}
w^{B}_{t, i} = w^{B}_{t-1, i}   p \left( \bx^B_{t, i} \big|  M\left( \rd_t, X^A_t \right) \right).
\label{eq:update}
\end{equation}

A straightforward but naive way of implementing the right-most factor in \eqref{eq:update} is to include the entire particle set $X^A_t$ in the message. Then
\begin{equation}
  p(\bx^B_{t, i} | \rd_t, X^A_t)= \sum_{j=1}^N  w^{A}_{t, j} p(\bx^B_{t, i}| \rd_t, \bx^{A}_{t, j}),
  \label{eq:naive}
\end{equation}
where $p(\bx^B | \rd, \bx^A)$ is the detection model, which we model as a normal distribution with variance $\Sigma$ (detection noise)
around a position corresponding to $\rd$
\begin{equation}
 \label{eq:model}
 p\left(\b x^B \big |  \rd, \b x^A\right) = \mathcal{N}\left ((x, y)^B; T(\rd; \b x^A), \Sigma \right),
\end{equation}
where $T(\cdot; \b x^A)$ transforms a relative position with respect to pose $\b x^A$ to an absolute position; please note that detections have no information about orientation.
%
This naive implementation has quadratic time-complexity (see Section~\ref{sec:CA}), which is problematic when resources are limited. 
For this reason, in \secref{sec:DC}, we compress the belief of $A$ in $M$, summarizing it with fewer representative points.



 \subsubsection{Reciprocal Sampling}\label{sec:sample}
\begin{figure}[tb] 
 \todo[inline]{Make the labels readable}
  \centering
	\begin{subfigure}{\linewidth}
		\includegraphics[width=\linewidth]{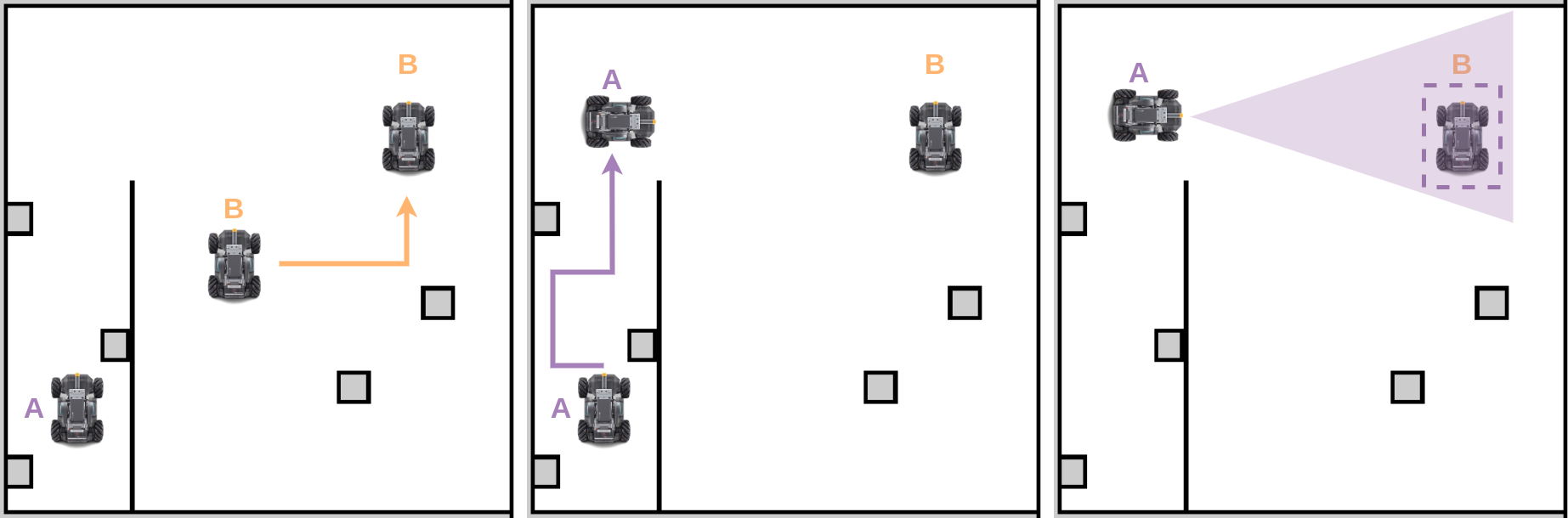}
		\caption{}
		\label{fig:detection}
	\end{subfigure}
         \begin{subfigure}{\linewidth}
		\includegraphics[width=\linewidth]{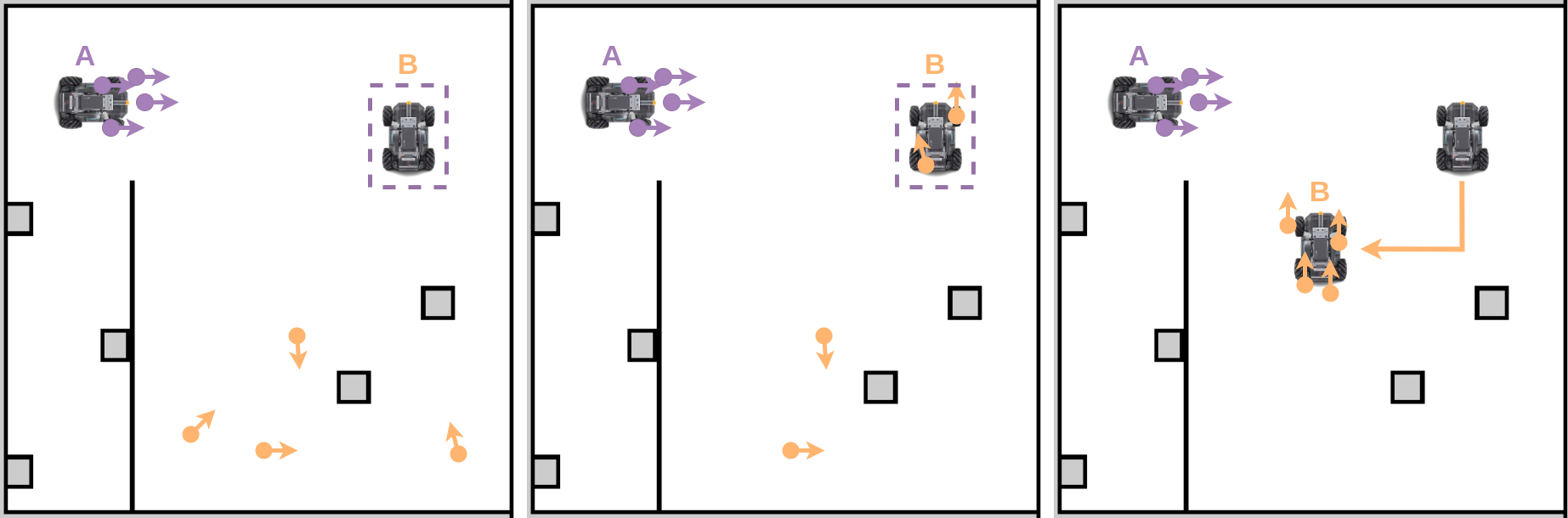}
		\caption{}
		\label{fig:recisampling}
	\end{subfigure}
  \caption{
  (a) An illustration of an experimental run up to the first detection event.
  (b) A run where robot $B$ is has no particles around its truth position at the time
  of detection (left), followed by reciprocal sampling (center) and successful localization (right).} 
  
  \label{fig:cmcl}
\end{figure}

To accelerate convergence, Prorok~\etalcite{prorok2012iros}, during the particle filter resampling step, propose sampling from the detection distribution (i.e., the right-most factor in \eqref{eq:update}) with probability $\alpha > 0$. In essence, particles in the detected robot's filter are replaced with particles suggested by the detecting robot, based on its belief and the estimated relative position. The reciprocal sampling procedure is illustrated in \figref{fig:recisampling}.

\subsection{Distribution Compression} \label{sec:DC}
We present our approach for near-linear time distribution compression, based on Compress++~\cite{shetty2022iclr}.
This compression method performs better than standard thinning algorithms, such as independent and identically distributed~(i.i.d.) sampling, which are not concise and have a large integration error~\cite{dwivedi2022iclr}. 
An important measure is the maximum mean discrepancy (MMD)~\cite{gretton2012jmlr}, which measures distance between two distribution or sample sets, 
as a difference between mean embedding of features. The MMD between two sample sets $X_1$ and $X_2$ is defined as
\begin{align}
\mathrm{MMD}(X_1, X_2) \doteq || \mathbb{E}_{X_1}[K(X_1)] - \mathbb{E}_{X_2}[K(X_2)] ||_{\mathcal{H}},
\end{align}
where $k(\cdot)$ is the reproducing kernel, such that 
$K(X)\doteq \left(k(x_i, x_j)\right)_{ij}$
 is a symmetric positive semi-definite matrix 
over all input points $x_i \in X$. Kernel thinning (KT) algorithms~\cite{dwivedi2021arxiv, dwivedi2022iclr} use a better than i.i.d., non-uniform randomness to thin a sample set. 
KT algorithms recursively partition the input into balanced coresets, by ensuring each pair of coresets minimizes the MMD. In the initial step, there are two empty coresets, and two samples, x and x', from the original sample set are chosen at random. The assignment of each of the samples to a coreset is designed to minimize   $\mathrm{MMD}(X_1 \cup{} \: \{x\}, X_2\cup{} \:  \{x'\})$. This step is repeated until all samples from the original set are assigned to one of the two coresets. 
\todo{say that it is iterative. what are $x$ and $x'$?
}
This yields a near-optimal thinning procedure, which compresses a set of points while providing error guarantees. 
 However, it suffers from quadratic or super-quadratic runtime. Shetty~\etalcite{shetty2022iclr} introduce Compress++, a meta-procedure for 
 speeding up thinning algorithms while suffering at most a factor of 4 in error. This root-thinning algorithm returns a subset of $\sqrt{N}$
 samples, with time complexity of $O(N\log^3N)$. 

 In our formulation, we first compute a set  
 \begin{equation}
 \label{eq:TX}
 \bar X_t^{AB} = \{T(\rd_t; \b x^A) | ( \b x^A, \cdot) \in X^A_t\}  \subset \mathbb{R}^2
 \end{equation}
 of \emph{position} samples for $B$ from the particles set $X^A$ (after resampling, i.e., when particles have uniform weights) and then compress it using Compress++  to $\tilde X_t^{AB} \subset  \bar X_t^{AB} $: this representative subset is then used in \eqref{eq:naive} and \eqref{eq:model} instead of the whole set.  
We adopt the idea of reciprocal sampling for our distribution compression, by first sampling uniformly from $\tilde X_t^{AB}$ and then sampling a pose using  \eqref{eq:model}:
\begin{equation*}
  \mu \sim \mathcal{U}(\tilde X_t^{AB}), \quad
  (x, y)_{t,i}^B \sim  \mathcal{N}(\mu, \Sigma),  \quad
  \theta_{t,i}^B \sim \mathcal{U}([0, 2\pi)).
  \label{eq:compress_sampling}
\end{equation*}

\subsection{Baselines} \label{sec:baselines}
As part of our contribution, we implement several baselines, including seminal works~\cite{fox2000ar,prorok2012iros} 
in collaborative localization. We provide an overview of these approaches, using common notations for clarity. 
\subsubsection{Density Estimation Tree}
Fox~\etalcite{fox2000ar} propose to use density estimation trees (DET)~\cite{omohundro1990nips, ram2011kddm} to transform the sample set 
into a piece-wise constant density function. 
First, $\mathrm{DET}^{AB}$ is constructed from $\bar X_t^{AB}$ and shared with $B$, who then updates its belief by querying it:
\begin{align}
  p\left(\b x_{t,i}^B \big | M\left(\rd_t, X_t^A \right) \right)  = \mathrm{DET}^{AB}((x, y)_{t,i}^B). 
  \label{eq:det}
\end{align}
While DET was suggested as a remedy for the nontrivial issues of establishing correspondence between two sample sets ($X^B$ and $\bar X_t^{AB}$) without an explicit detection model, it can be used to compress the distribution by limiting the size $T$ of the tree. For this method, we do not perform reciprocal sampling because the implementation detailed in the paper does not include it.

\subsubsection{Divide-and-Conquer Clustering}
Prorok~\etalcite{prorok2012iros}  propose a non-iterative, order-independent, non-parametric clustering
inspired by multidimensional binary trees~\cite{bentley1975cacm}.

\noindent
The particles of $A$ are clustered into $K$ cluster abstractions, 
which include the centroid $\b c_{t,k}^A$, weight $w^A_k$, detection mean  $\mu^A_k$ and detection variance $\Sigma^A_k$. 
The belief is updated as
\begin{scriptsize}%
\begin{equation*}
  p\left(\b x_{t,i}^B \Big | M\left(\rd_t, X_t^A \right) \right) = \sum_{k=1}^K w^A_k \mathcal{N}(T^{-1}\left((x, y)_{t,i}^B; \b c_{t,k}^A\right); \mu^A_k, \Sigma^A_k+\Sigma),
  \label{eq:prorok}
\end{equation*}
\end{scriptsize}%
where $T^{-1}$ transforms absolute to relative positions, the multivariate normal distribution is represented in relative polar coordinates,
and $\Sigma$ captures the detection noise.

\subsubsection{K-means Clustering}
K-means clustering~\cite{Hartigan1979ap} is a commonly-used, low-cost iterative clustering method. 
It is mentioned by Prorok~\etalcite{prorok2012iros} in the context of collaborative localization, where
the authors dismiss it as too sensitive to the initial cluster assignment, but no comparison is reported.

\noindent
We introduce a belief compression based on K-means clustering: we compute $K$ clusters of $\bar X_t^{AB}$ with centroids $c_{t,k}^A$; for each cluster, we compute 
its intra-cluster variance $\Sigma^A_k$ and total weight $w^A_k$. The belief is then updated as
\begin{footnotesize}%
\begin{equation}
 \label{eq:kmeans}
  p\left(\b x_{t,i}^B \big | M\left(\rd_t, X_t^A \right) \right) = \sum_{k=1}^K w^A_k \mathcal{N}\left((x, y)_{t,i}^B; c_{t,j}^A, \Sigma^A_k \ + \Sigma \right) 
\end{equation}
  \end{footnotesize}%
%
%
By clustering $\bar X_t^{AB}$ instead of $X^A$, we reduce the amount of information to broadcast compared to 
the approach suggested by Prorok~\etalcite{prorok2012iros}. 
For the reciprocal sampling strategy, we sample from the mixture of normal distributions of \eqref{eq:kmeans}.

\subsubsection{Standard Thinning}
The most common approach to reduce the number of samples is i.i.d. sampling, where $K$ samples are randomly picked from a set of size $N$. 
This can be used to reduce $\bar X^{AB}$, share it and then apply \eqref{eq:naive}. In our implementation, we also include a reciprocal sampling step similar to the one preformed for our Compress++ approach.

\section{Complexity Analysis} \label{sec:CA}

The complexity of all approaches is presented in \tabref{tab:complexity}. 
We discuss the complexity of 3 components: compression, communication and fusion. 
\todo{Mention these 3 keywords in Intro and/or Approach}
These time and space complexities become significant in systems with many robots or when the computational and communication resources are constrained. 

\begin{table}[htb]
  \caption{Algorithm Complexity. $N$ is the number of particles. $K$ is the number of clusters for Prorok et al. and K-means, and the number of points selected by standard thinning.}
   \centering
   \resizebox{\columnwidth}{!}{
 \begin{tabular}{lcccccc}\toprule
 Method & Naive &Std. Thinning & Fox et al. & Prorok et al.& K-means & Compress++ \\ \midrule
 Compression & $O(N)$ & $O(K)$ & $O(HDN\log N)$ &  $O(NK)$ & $O(NKL)$ & $O(N\log^3N)$     \\
 Communication & $O(N)$ & $O(K)$ & $O(N)$ &  $O(K)$ & $O(K)$ & $\sqrt{N}$     \\
Fusion & $O(N^2)$ & $O(NK)$ & $O(N\log N)$ & $O(NK)$ & $O(NK)$ & $O(N\sqrt{N})$\\
  \bottomrule
 \end{tabular}
 }
 \label{tab:complexity}
\end{table}

\subsection{Compression}
The naive implementation for belief exchange require just $O(N)$ complexity for compression, but results
in high communication cost and computational cost during fusion. The construction of a DET involves leave-one-out cross-validation, resulting in a $O(HDN \log N)$ complexity for $N$ data samples with $D$ features and $H$ tries. 
For the non-iterative clustering method suggested by Prorok~\etalcite{prorok2012iros}, the complexity 
for 
$K$ clusters is $O(NK)$.  The complexity of K-means clustering is $O(NKL)$, where
$L$ in the number of iterations. Even though K-means clustering is an iterative approach, for $L=5$ this compression strategy is comparable to Prorok et al.'s approach, and in practice runs faster (see \secref{sec:runtime}). 
The complexity of the standard thinning algorithm is $O(K)$, where $K$ is the size of the reduced sample set.
To compress a state with $N$ particles, our Compress++ approach has a time complexity of $O(N\log^3N)$.
\subsection{Communication}
When multiple robots communicate on the same network, the total bandwidth required needs to be considered.
The naive approach communicates the robot's belief by broadcasting all particle in the filter, which takes $O(N)$ space.
To reduce the amount of information we broadcast, it is necessary to compress the belief. When the belief is represented as a 
DET~(\cite{fox2000ar}), the information sent is in the order of $O(N)$, with constant coefficient that represents the space required for the bookkeeping of a single tree node. When the size of the tree is limited to $T$ nodes, the cost is reduced to $O(T)$. Both Prorok et al.'s and K-means clustering use cluster representatives to summarize the belief, resulting in $O(K)$ space for $K$ clusters. Similarly, for standard thinning, we only broadcast the reduced set of $K$ particles. 
Using our Compress++ approach, the coreset representation reduces the communication cost to $O(\sqrt{N})$ ($K=\sqrt{N}$ in this case).

\subsection{Fusion}
Following \eqref{eq:naive}, the complexity of updating the belief is $O(N^2)$ for the naive implementation. 
Fox~\etalcite{fox2000ar} requires $O(N \log N)$, since it involves querying a tree.
The update step, combined with the reciprocal sampling, detailed by Prorok~\etalcite{prorok2011iros} requires $O(NK)$ time; the same for K-means and std. thinning.
Our Compress++ approach has also a $O(N\sqrt{N})$ complexity when updating the belief ($K=\sqrt{N}$ in this case).

\section{Experimental Setup}    \label{sec:setup}
We restrict the study to the case of two robots, although all presented methods generalize to larger multi-robot systems. In our experiments, robot $B$ is delocalized at the time when it is first detected by robot $A$; we explore the contribution of information exchange to the localization performance of robot $B$.

\subsection{Robots}

\begin{figure}[htb] 
  \centering
  \includegraphics[trim={0.5cm 0.5cm 1.75cm 0.6cm},clip, width=0.6\linewidth]{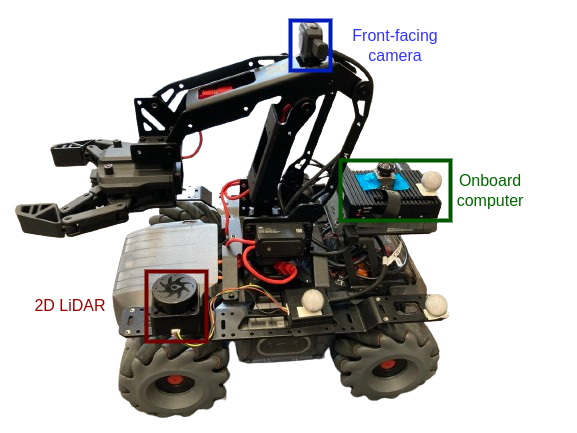}  
  \caption{The robotic platform used in the evaluation.} 
  \label{fig:robomaster}
\end{figure}

The platform we use in the evaluation is the DJI RoboMaster EP, a commercially available ground robot, with omnidirectional drive, and a size of 
$32 \times 24 \times 27$\,cm.
Depicted in \figref{fig:robomaster}, 
the robot has a front-facing CMOS HD camera with a \ang{102} horizontal FoV and a YDLIDAR Tmini Pro 2D LiDAR with a range of \SI{12}{\m}, \ang{360} FoV, and resolution of \ang{0.54} at \SI{6}{\Hz}.
The onboard robot firmware includes an ML model for detecting other RoboMasters in the camera stream. The detector runs at about \SI{5}{\Hz} and returns a list of bounding boxes in image space, from which, using calibrated homography, 
we reconstruct the relative horizontal position of detected robots $\rd = (r, \theta)$, with a zero-mean gaussian error ($\sigma_r \approx 0.05 r, \sigma_{\theta} \approx  \SI{0.03}{\radian}$) that depends on the range.
The robots carry a Single Computer Board (Khadas VIM4) that runs ROS2 drivers for the robot platform\footnote{\url{https://github.com/jeguzzi/robomaster_ros}} and LiDAR.
 
The same platform is available in simulation\footnote{\url{https://github.com/jeguzzi/robomaster_sim}} (CoppeliaSim~\cite{rohmer2013iros}), where we replicate the same sensing error model and run the same ROS2 driver.

\subsection{Environments}  
\begin{figure}[htb]
	\centering
		\includegraphics[width=\linewidth]{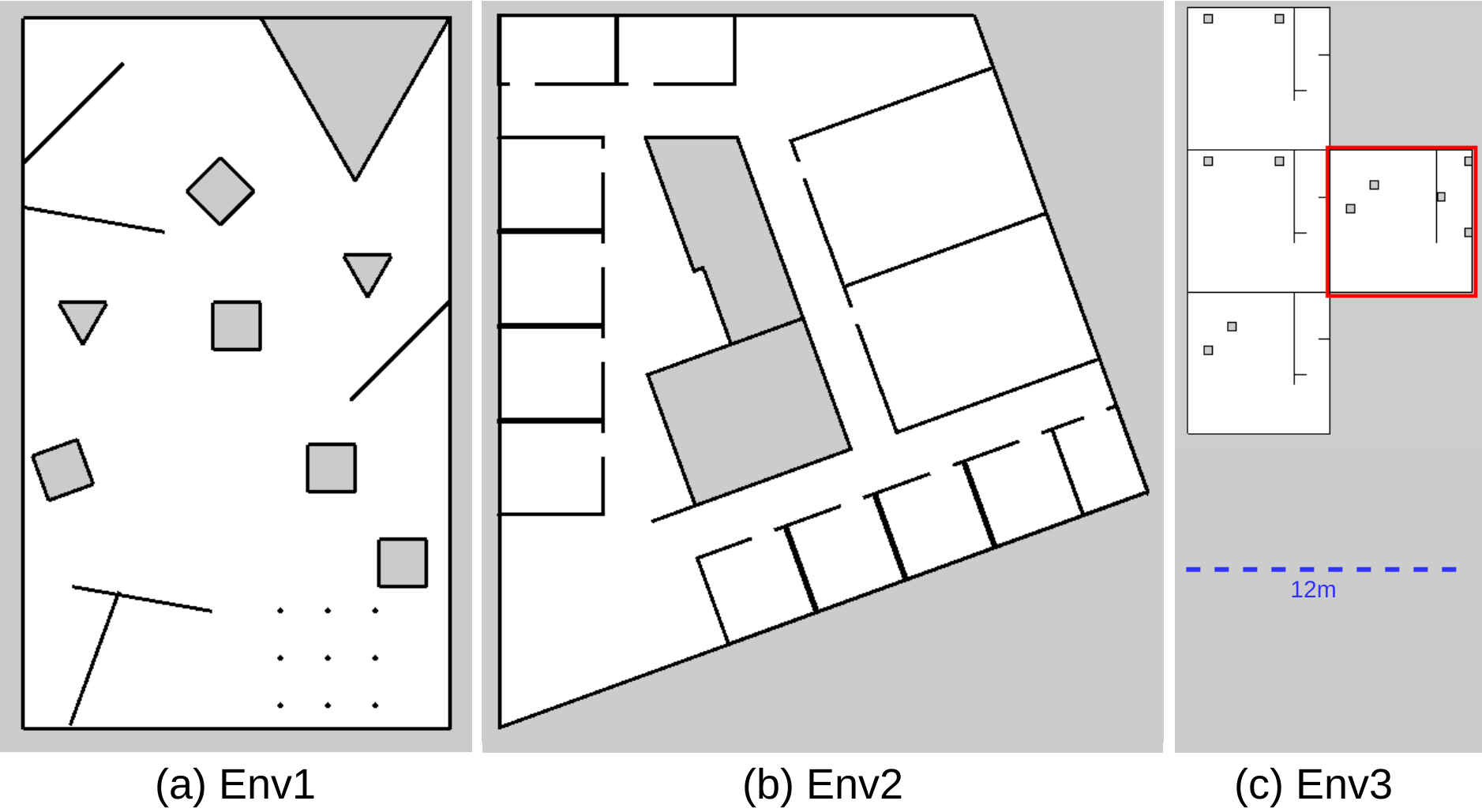}
	\caption{Environments have varying degree of geometric symmetry and feature richness. The area highlighted in red was reconstructed in our lab for real-world evaluation. The three maps are to scale; the LiDAR range, 12\,m, is marked in blue.}
	\label{fig:maps} 
\end{figure}

In simulation, we design three different environments~(\figref{fig:maps}), with a varying degree of geometric symmetry and feature richness.
Environment 1 is specifically designed to be feature-sparse, with multiple areas that challenge the limited range (\SI{12}{m}) of our LiDAR.  Environment 2 is modeled after 
a floor of our building. Environment 3 is designed to have a large degree of symmetry.
Environments 1 and 2 have a free area of \SI{500}{\metre\squared} where experiments are executed with 10000 particles. 
Environment 3 is smaller, with a free area of \SI{140}{\metre\squared} where experiments are executed with 2000 particles.
To conduct real-world evaluation, we reconstruct the layout of the left room in environment 3 in our lab, which features a motion tracker system to collect ground truth information.


\subsection{Scenario}
We focus our experimental scenario on the impact of collaborative localization, therefore we make sure that each experimental run includes multiple detection events. 
We randomly choose  start and goal poses of robot $A$  and compute a shortest-path trajectory to follow.
We then choose the start pose for robot $B$ such that it would be seen by robot $A$ at some point along its trajectory, and a random goal pose. In this scenario, robot $B$ is delocalized at the time of the first detection by robot $A$, as illustrated in \figref{fig:detection}.

\subsection{Metrics}
Three metrics were considered for the evaluation - the success rate, absolute trajectory error~(ATE) after convergence and convergence time. 
We define convergence as the time when the estimate pose is within \SI{0.3}{m} radius of the ground truth pose, and the orientation is within 
\SI{0.3}{\radian}. After convergence, the tracked pose must not diverge for an accumulated 5\% of the remaining sequence.
A localization run is successful if convergence is achieved in the first 90\% of the sequence time and the tracked pose 
does not diverge. Each sequence is evaluated multiple times to account for the inherent stochasticity of the MCL framework.

\subsection{Procedure and parameters}

We first record odometry, LiDAR scans, detections and ground truth poses for each robots during all experimental runs. 
In total, we recorded 19 runs for the 3 environments in simulation and 17 runs for Env3 in real-world.
Then, we evaluate each method presented in \secref{sec:approach} on the same runs, using the parameters reported in \tabref{tab:parameters} for the common MCL part. 
 Overall, 112 evaluations were run for each of the 7 methods.
 
\begin{table}[htb]
  \caption{Common MCL parameters} 
   \centering
   \resizebox{\columnwidth}{!}{
 \begin{tabular}{ccccccc}\toprule
$\sigma_{\mathrm{odom}}$  &  $\sigma_{\mathrm{obs}}$ & $r_{\mathrm{max}}$ & $ \alpha $ &  $\delta_{xy}$ & $\delta_{\theta}$ \\ \midrule
 $(0.05, 0.05, 0.05)$ & \SI{0.5}{m} & \SI{12.0}{m} & 0.06 &   \SI{0.05}{m} & \SI{0.05}{\radian}  \\
  \bottomrule
 \end{tabular}
 }
 \label{tab:parameters}
\end{table}
 

The Compress++ algorithm reduces the sample set first to the closest power of 4, and then perform root-thinning, 
which resulted in 64 representative samples from a set of 10000 particles, and 32 representative samples for a sample set of 2000 particles. For Fox at al.'s approach, we ensured that we have no more than 20 leaves in the DET. 
Prorok~\etalcite{prorok2011iros}  explored different cluster numbers, between 1 and 32, and reported no significant change in performance for any $K > 1$. Therefore, for both Prorok et al.'s and the K-means approaches, we set $K=8$. The reciprocal sampling ratio was set to $\alpha=0.06$, as suggested by Prorok~\etalcite{prorok2011iros}. 

\section{Experimental Evaluation} \label{sec:exp}

%
%
We conducted a thorough evaluation of the different approaches to cooperative localization. We present our experiments to show the capabilities of our method. 
The results 
support the claims that our proposed approach
 (i) improves collaborative localization, 
 (ii) decreases the required bandwidth, 
 (iii) reduces the computational load,
 (iv) runs online on an onboard computer.

\subsection{Collaborative localization}

We compare our approach against the different baselines. We focus on the impact of belief exchange between the somewhat-localized robot $A$ and the delocalized robot $B$, on the localization performance of robot $B$, for the different approaches.

\begin{figure}[tb]  
  \centering 
  \includegraphics[width=\linewidth]{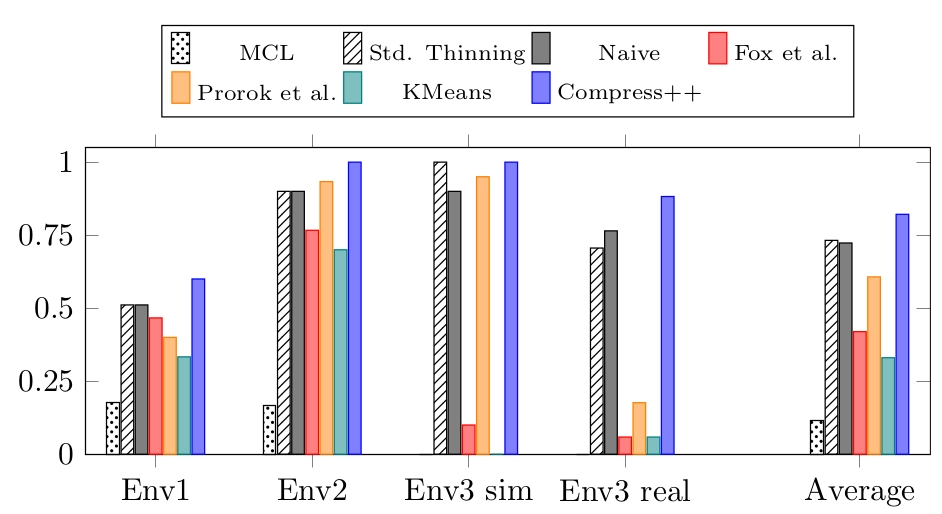}  
  
  \caption{The success rate of all methods for each of the environments for robot $B$.}
  \label{fig:successrate}
\end{figure}

\begin{table}[tb] 
  \caption{Baseline comparison of global localization performance for robot $B$. We report success rate, convergence time in seconds (top) and ATE in $[\mathrm{rad}/m]$ format (bottom). ATE is not reported when all runs resulted in failure.}
  \centering 
  \resizebox{\columnwidth}{!}{
\begin{tabular}{lccccc}
  \toprule
  Method       & Env1 & Env2 & Env3 (sim) & Env3 (real) & Average\\  \midrule
MCL 	 & 17.8\% (11.2)	 & 16.7\% (7.8)	 & 0.0\% (-)	 & 0.0\% (-)	 & 11.6\% (6.6)\\
Std. Thinning 	 & 51.1\% (32.4)	 & 90.0\% (24.5)	 & \textbf{100.0\%} (45.6)	 & 70.6\% (43.8)	 & 73.2\% (34.4)\\
Naive 	 & 51.1\% (29.7)	 & 90.0\% (24.6)	 & 90.0\% (53.1)	 & 76.5\% (38.2)	 & 72.3\% (33.8)\\
Fox et al.	 & 46.7\% (34.8)	 & 76.7\% (17.4)	 & 10.0\% (91.1)	 & 5.9\% (39.5)	 & 42.0\% (40.9)\\
Prorok et al. 	 & 40.0\% (43.3)	 & 93.3\% (18.3)	 & 95.0\% (41.0)	 & 17.6\% (27.5)	 & 60.7\% (33.8)\\
K-means 	 & 33.3\% (51.9)	 & 70.0\% (16.5)	 & 0.0\% (-)	 & 5.9\% (39.9)	 & 33.0\% (31.3)\\
Compress++ 	 & \textbf{60.0\%} (34.2)	 & \textbf{100.0\%} (23.1)	 & \textbf{100.0\%} (47.9)	 & \textbf{88.2\%} (41.2)	 & \textbf{82.1\%} (34.7)\\
\midrule
MCL 	 & 0.032/0.198	 & 0.005/0.175	 & -/-	 & -/-	 & 0.014/0.127\\
Std. Thinning 	 & 0.026/0.199	 & 0.022/0.103	 & 0.031/0.062	 & 0.062/0.153	 & 0.031/0.142\\
Naive 	 & 0.026/0.201	 & 0.021/0.106	 & 0.023/0.071	 & 0.060/0.186	 & 0.029/0.150\\
Fox et al. 	 & 0.029/0.217	 & 0.045/0.142	 & 0.048/0.185	 & 0.181/0.117	 & 0.060/0.176\\
Prorok et al. 	 & 0.034/0.199	 & 0.036/0.121	 & 0.068/0.099	 & 0.090/0.114	 & 0.049/0.147\\
K-means 	 & 0.030/0.193	 & 0.033/0.163	 & -/-	 & 0.156/0.171	 & 0.045/0.147\\
Compress++ 	 & 0.024/0.214	 & 0.018/0.108	 & 0.026/0.062	 & 0.060/0.154	 & 0.028/0.149\\
    \bottomrule
\end{tabular}}
\label{tab:localization} 
\end{table}

As can be seen from \figref{fig:successrate} and \tabref{tab:localization}, the results highlight the importance of reciprocal sampling. The seminal work by Fox \etalcite{fox2000ar} does not include a reciprocal sampling step and  performs poorly in many sequences. To support this claim, we evaluated the performance of the naive implementation with no reciprocal sampling, which result in a dramatic drop of performance, with success rates of (20\%, 10\%, 40\%) in the 3 simulated environments respectively. As illustrate in \figref{fig:recisampling}, reciprocal sampling is particularly crucial when robot $B$ is delocalized and few-to-none particles are present around its true location, as reweighting particles in \eqref{eq:update} has minimal impact: it can only encourage particles that are in the vicinity of the detection position, but not propose new hypotheses to robot $B$. In contrast, reciprocal sampling allows robot $A$ to enrich robot $B$'s particle filter with new samples.

Another interesting insight comes from the difference in performance in different environments. Due to the sparsity of environment 1, robot $A$'s localization is less accurate prior to the detection event. For the real-world environment, localization is challenging due to discrepancies between the map and the constructed maze. Additionally, the LiDAR is partially occluded by the arm, reducing the FoV to \ang{300}. For these reasons, the localization for robot $B$ is challenging even after integrating $A$'s belief, as indicated by the higher ATE across all methods~(\tabref{tab:localization}) in these two environments. Methods that perform well on the second and third~(simulated) environment, like standard thinning and the approach of Prorok et al., suffer a significant loss in performance in the first environment, as well as in the real-world experiments. Our approach, Compress++, remains a top-performer in all 3 environments, including in the real-world, supporting our first claim. We provide a demonstration of our approach in a real environment in the attached video.

\begin{figure}[tb]      
  \centering
  \includegraphics[width=\linewidth]{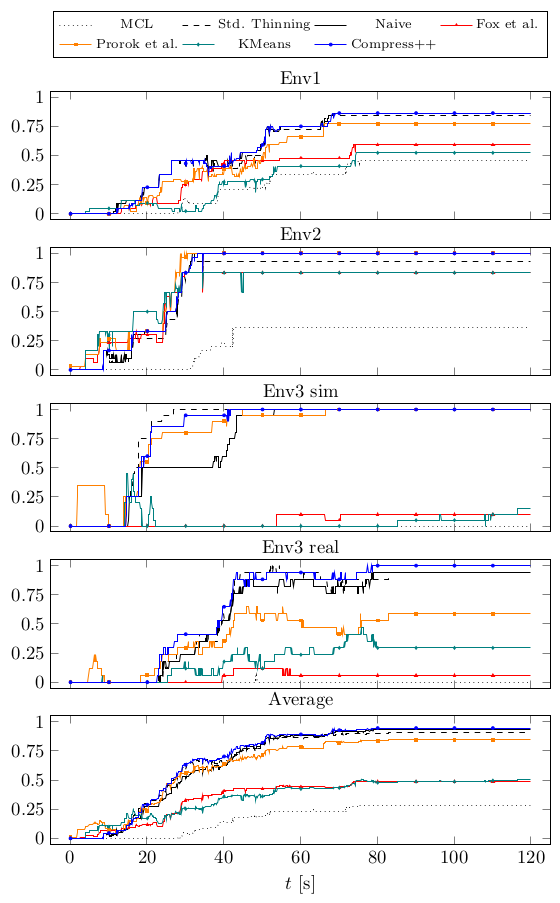}  
  \caption{The fraction of runs whose current pose estimation of robot $B$ is below convergence threshold. The time $t=0$ is synchronized by the arrival of the first detection message from robot $A$.}
  \label{fig:conv} 
\end{figure}

In \figref{fig:conv}, we visualize convergence as a function of time, where all sequences are synchronized such that $t=0$ when the first detection message is received by robot $B$. For every time step, we report for what fraction of the runs the current pose estimation is below the convergence threshold. 
While in  \figref{fig:successrate} and \tabref{tab:localization} success requires that pose estimation remains within a convergence threshold from the moment of convergence, in \figref{fig:conv}, we only consider whether a pose estimation is close enough to the ground truth at a given time.
Therefore, \figref{fig:conv} differs from results about success, 
yet shows a similar ranking between the methods, with Compress++ converging with the highest reliability. We note that while Prorok et al.'s approach converges well in the long run, it suffers particularly from instability as it tends to converge fast and then diverge, as seen in the drop around $t=\SI{5}{s}$ and $t=\SI{60}{s}$ in Env3 (real). 
Similar differences  between convergence and success rate can be seen for all methods but to a lesser degree.

\subsection{Bandwidth requirements}
\todo{We miss Std Thinning}
We go through the implementation of all methods to compute how much bandwidth they require.
For the naive approach, where the entire distribution is exchanged, the size of a message is $12N=120$ kB as each particle is represented by 3 floating point numbers, 
assuming 4-byte representation for floats.
For the K-means approach, we require 6 floating point numbers to encode each cluster, i.e., a message size of $24K=192$ bytes for $K=8$ clusters. 
For Prorok et al.'s approach, we need 8 floating point numbers to 
describe each cluster abstraction, giving a message of 256 bytes. 
For Fox et al.'s, the whole DET is sent; since each node requires some 
bookkeeping (50 bytes per node, a conservative estimation), it requires $50T=1$ kB for the entire tree when $T=20$.  For standard thinning, we require $3K=192$ bytes for $K=64$ sampled particles. 
For Compress++, representative points have equal weight and are represented by 2 floats; as it first reduces the set to the nearest power of 4, and then takes the root, it
requires 512 bytes for 10000 particles and 256 bytes for 2000 particles. 
Therefore our approach drastically reduces the bandwidth requirement compared to the naive approach (second claim), achieving a comparable compression rate as the other methods but maintaining a larger localization performance.

\subsection{Runtime cost} \label{sec:runtime}

As discussed in \secref{sec:CA}, our approach reduces the overall time-complexity compared to the naive approach: on one side, it increases the time-complexity of compression for robot $A$ by factor $\log^3 N$, 
on the other side, it decreases the larger time-complexity for robot $B$ by a more significant factor $\sqrt{N}$, supporting our third claim.


The runtime cost for all collaborative localization approaches is presented in \tabref{tab:runtime}.
We benchmarked the approaches using 10000 particles. Since the output of Compress++ is 64 representative
points, we select a comparable $K=64$ for Prorok et al., std. thinning, and K-means, and constrain the 
DET construction to result in about 64 nodes. As expected, the naive approach requires the longest time to perform the belief fusion.

\begin{table}[htb] 
  \caption{Runtime cost  in milliseconds for one update step of filters with 10000 particles.}
  \centering 
  \resizebox{\columnwidth}{!}{
\begin{tabular}{llcccccc}\toprule 
HW & Method    & Std. Thinning    &  Naive  & Fox et al. & Prorok et al.& K-means & Compress++ \\  \midrule
Laptop & Compression ($A$) & 0.1 & 0.1 & 5.4 & 20.8& 1.6 & 32.2\\
 & Fusion ($B$)  & 2.6 & 286.1 & 0.4 &  1.0 & 1.5 &2.5\\
\bottomrule 
Khadas & Compression ($A$) &  0.2 & 0.3  & 10.9 & 51 & 8.8 & 85 \\
& Fusion ($B$) & 9.9 & 1191.7 & 0.6 & 4.0 & 5.9 & 9.9 \\
\bottomrule 
\end{tabular}
}
\todo[inline]{Let's keep the same method  order in all tables and plots}
\label{tab:runtime}   
\end{table}

We evaluated the performance on two platform, a laptop with Intel Core i9 processors~(16 cores), and Khadas VIM4 board with ARM Cortex processors~(8 cores).
Even though K-means clustering has a slightly higher theoretical complexity,
 for $L=5$ this compression strategy is comparable to approach of Prorok et al., and in practice runs 12 times faster on the laptop and 5.8 times faster on the embedded platform.  
We attribute this performance gap to the fact that K-means is much easier to parallelize. While all other methods are implemented in C++ and are optimized for multi-core execution, our method utilizes a
open-source Python implementation of Compress++.\footnote{https://github.com/microsoft/goodpoints/tree/main} Nonetheless, its runtime for compression is on-par with the approach of Prorok et al. and takes 
less than \SI{10}{\ms} to integrate the belief, supporting our fourth claim that it can run in real-time onboard.
Additionally, the attached video shows how our localization method runs online, onboard of the robots.

\subsection{Clustering} 

\begin{figure}[htb] 
  \centering
	\begin{subfigure}{0.49\linewidth}
    \includegraphics[width=\linewidth]{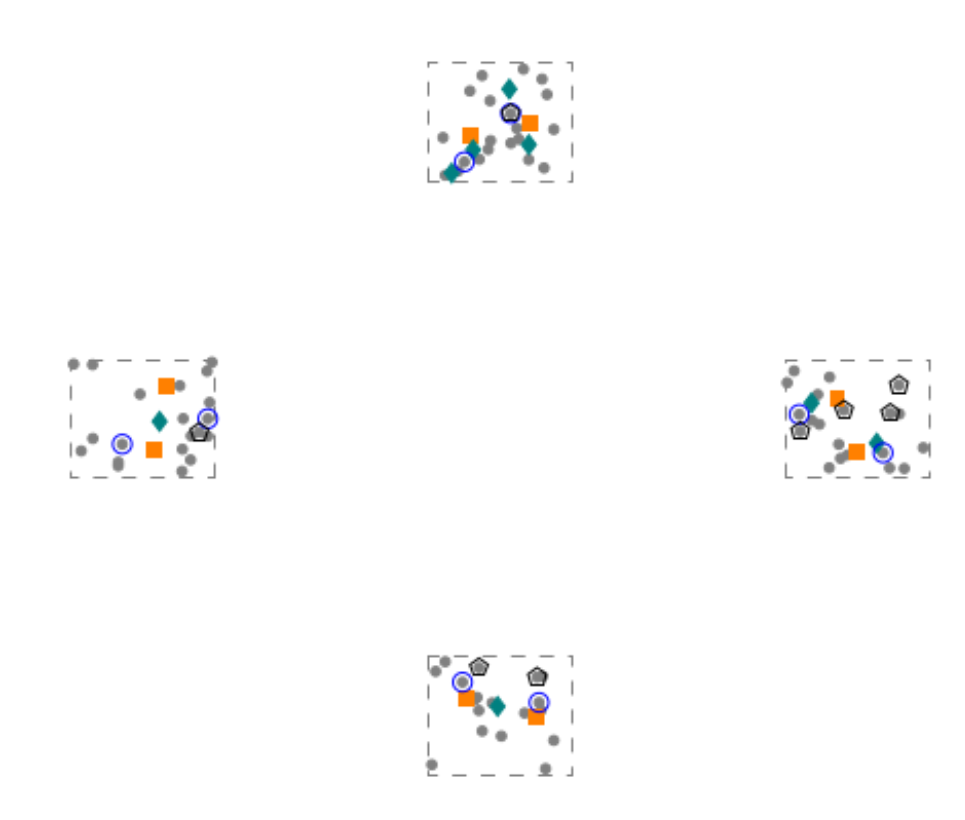}  
		\caption{}
		\label{fig:clustering2}
	\end{subfigure}
	\unskip\ \vrule
         \begin{subfigure}{0.49\linewidth}
    \includegraphics[width=\linewidth]{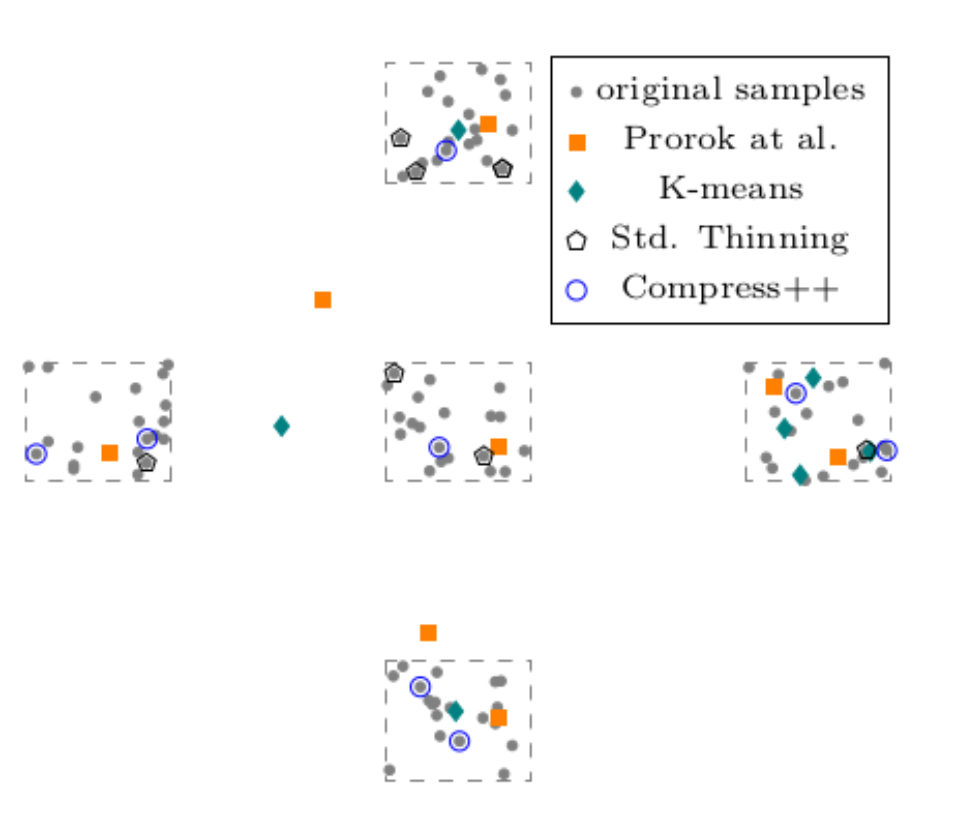} 
		\caption{}
		\label{fig:clustering1}
	\end{subfigure}
   \caption{The behavior of different distribution compression methods on different data points formations.}
   \todo[inline]{It would be better to generate points with less white space in-between, i.e., either increase the rectangles size or move them closer. The figure would regenerate automatically.}
   \label{fig:prorokfail} 
\end{figure}

We explore an artificial yet interesting case of compression in the presence of symmetry, as 
would occur when a robot is not yet localized but maintains several hypothesis, e.g., due to geometric symmetry of the environment. We generated 20 points each from 4 evenly spaced regions in a diamond formation (\figref{fig:clustering2}) and another 20 points from a region at the center (\figref{fig:clustering1}). We then apply different methods to extracts 8 representative points. 

This test case reveals a weakness of the divide-and-conquer clustering algorithm proposed by Prorok~\etalcite{prorok2012iros}: as shown in \figref{fig:clustering1}, the algorithm struggles to divide the particle set into well-defined clusters, likely due to the division strategy, which separates clusters along the axis of highest variance. The algorithm also fails when an even number of clusters are places along a circle with an additional one in the middle.

K-means clustering also fails to perfectly segment the clusters on this type of points distribution. The standard thinning approach, i.e., sampling 8 points at random, is naturally less expressive than other density estimation approaches, yet, in this case, it represents 4 out of the 5 clusters. Compress++ succeeds to extract samples from each cluster.




\section{Conclusion}
\label{sec:conclusion}
In this paper, we presented a novel approach to resource-aware collaborative global localization. We also provided a detailed overview of different distribution compression approaches, as well as a C++/ROS2 implementation. Additionally, we conducted a thorough complexity analysis and bench-marking for the various methods, which we also open-source for the benefit of the community. In the future, we would like to extend the experiments to larger groups of  robots, as well as to expand the real-world experimental setup beyond a single room. 



\bibliographystyle{plain_abbrv}

\bibliography{glorified,new}

\begin{thebibliography}{10}

\bibitem{atanasov2015ijrr}
N.~Atanasov, M.~Zhu, K.~Daniilidis, and G.J. Pappas.
\newblock {Localization from semantic observations via the matrix permanent}.
\newblock {\em Intl.~Journal~of Robotics Research (IJRR)}, 35(1-3):73--99,
  2016.

\bibitem{barea2007itssa}
R.~Barea, E.~L{\'o}pez, L.M. Bergasa, S.~{\'A}lvarez, and M.~Oca{\~n}a.
\newblock {Collaborative multi-robot Monte Carlo localization in assistant
  robots}.
\newblock {\em International Transactions on Systems Science and Applications},
  3(3):227--237, 2007.

\bibitem{bentley1975cacm}
J.L. Bentley.
\newblock {Multidimensional binary search trees used for associative
  searching}.
\newblock {\em Communications of the ACM}, 18(9):509--517, 1975.

\bibitem{bochkovskiy2020arxiv}
A.~Bochkovskiy, C.Y. Wang, and H.Y.M. Liao.
\newblock {YOLOv4: Optimal Speed and Accuracy of Object Detection}.
\newblock {\em arXiv preprint}, 2004.10934, 2020.

\bibitem{cadena2016tro}
C.~Cadena, L.~Carlone, H.~Carrillo, Y.~Latif, D.~Scaramuzza, J.~Neira, I.~Reid,
  and J.~Leonard.
\newblock {Past, Present, and Future of Simultaneous Localization And Mapping:
  Towards the Robust-Perception Age}.
\newblock {\em IEEE Trans.~on Robotics (TRO)}, 32:1309--1332, 2016.

\bibitem{cheng2020cvpr}
B.~Cheng, M.D. Collins, Y.~Zhu, T.~Liu, T.S. Huang, H.~Adam, and L.C. Chen.
\newblock {Panoptic-DeepLab: A Simple, Strong, and Fast Baseline for Bottom-Up
  Panoptic Segmentation}.
\newblock In {\em Proc.~of the IEEE/CVF Conf.~on Computer Vision and Pattern
  Recognition (CVPR)}, 2020.

\bibitem{cui2021iros}
L.~Cui, C.~Rong, J.~Huang, A.~Rosendo, and L.~Kneip.
\newblock {Monte-Carlo Localization in Underground Parking Lots Using Parking
  Slot Numbers}.
\newblock In {\em Proc.~of the IEEE/RSJ Intl.~Conf.~on Intelligent Robots and
  Systems (IROS)}, 2021.

\bibitem{dellaert1999icra}
F.~Dellaert, D.~Fox, W.~Burgard, and S.~Thrun.
\newblock Monte carlo localization for mobile robots.
\newblock In {\em Proc.~of the IEEE Intl.~Conf.~on Robotics \& Automation
  (ICRA)}, 1999.

\bibitem{dwivedi2021arxiv}
R.~Dwivedi and L.~Mackey.
\newblock {Kernel Thinning}.
\newblock {\em arXiv preprint arXiv:2105.05842}, 2021.

\bibitem{dwivedi2022iclr}
R.~Dwivedi and L.~Mackey.
\newblock {Generalized Kernel Thinning}.
\newblock In {\em International Conference on Learning Representations}, 2022.

\bibitem{fox2000ar}
D.~Fox, W.~Burgard, H.~Kruppa, and S.~Thrun.
\newblock {A Probabilistic Approach to Collaborative Multi-robot Localization}.
\newblock {\em Autonomous Robots}, 8:325--344, 2000.

\bibitem{fox1999jair}
D.~Fox, W.~Burgard, and S.~Thrun.
\newblock {Markov localization for mobile robots in dynamic environments}.
\newblock {\em Journal of Artificial Intelligence Research (JAIR)},
  11:391--427, 1999.

\bibitem{gretton2012jmlr}
A.~Gretton, K.M. Borgwardt, M.J. Rasch, B.~Sch{\"o}lkopf, and A.~Smola.
\newblock {A kernel two-sample test}.
\newblock {\em The Journal of Machine Learning Research}, 13(1):723--773, 2012.

\bibitem{Hartigan1979ap}
J.A. Hartigan and M.A. Wong.
\newblock {A k-means clustering algorithm}.
\newblock {\em JSTOR: Applied Statistics}, 28(1):100--108, 1979.

\bibitem{he2017iccv-mr}
K.~He, G.~Gkioxari, P.~Doll{\'a}r, and R.~Girshick.
\newblock {Mask R-CNN}.
\newblock In {\em Proc.~of the IEEE Intl.~Conf.~on Computer Vision (ICCV)},
  2017.

\bibitem{hendrikx2021icra}
R.~Hendrikx, P.~Pauwels, E.~Torta, H.~Bruyninckx, and M.~van~de Molengraft.
\newblock {Connecting Semantic Building Information Models and Robotics: An
  application to 2D LiDAR-based localization}.
\newblock In {\em Proc.~of the IEEE Intl.~Conf.~on Robotics \& Automation
  (ICRA)}, 2021.

\bibitem{howard2002iros}
A.~Howard, M.J. Matark, and G.S. Sukhatme.
\newblock {Localization for mobile robot teams using maximum likelihood
  estimation}.
\newblock In {\em Proc.~of the IEEE/RSJ Intl.~Conf.~on Intelligent Robots and
  Systems (IROS)}, 2002.

\bibitem{ito2014icra}
S.~Ito, F.~Endres, M.~Kuderer, G.~Tipaldi, C.~Stachniss, and W.~Burgard.
\newblock {W-RGB-D: Floor-Plan-Based Indoor Global Localization Using a Depth
  Camera and WiFi}.
\newblock In {\em Proc.~of the IEEE Intl.~Conf.~on Robotics \& Automation
  (ICRA)}, 2014.

\bibitem{joho2009icra}
D.~Joho, C.~Plagemann, and W.~Burgard.
\newblock {Modeling RFID signal strength and tag detection for localization and
  mapping}.
\newblock In {\em Proc.~of the IEEE Intl.~Conf.~on Robotics \& Automation
  (ICRA)}, 2009.

\bibitem{martinelli2005icra}
A.~Martinelli, F.~Pont, and R.~Siegwart.
\newblock {Multi-robot Localization using Relative Observations}.
\newblock In {\em Proc.~of the IEEE Intl.~Conf.~on Robotics \& Automation
  (ICRA)}, 2005.

\bibitem{moravec1989sdsr}
H.P. Moravec.
\newblock {Sensor Fusion in Certainty Grids for Mobile Robots}.
\newblock In {\em Sensor Devices and Systems for Robotics (SDSR)}, 1989.

\bibitem{nerurkar2010iros}
E.D. Nerurkar and S.I. Roumeliotis.
\newblock {Asynchronous multi-centralized cooperative localization}.
\newblock In {\em Proc.~of the IEEE/RSJ Intl.~Conf.~on Intelligent Robots and
  Systems (IROS)}, 2010.

\bibitem{nerurkar2009icra}
E.D. Nerurkar, S.I. Roumeliotis, and A.~Martinelli.
\newblock {Distributed maximum a posteriori estimation for multi-robot
  cooperative localization}.
\newblock In {\em Proc.~of the IEEE Intl.~Conf.~on Robotics \& Automation
  (ICRA)}, 2009.

\bibitem{omohundro1990nips}
S.~Omohundro.
\newblock {Bumptrees for efficient function, constraint and classification
  learning}.
\newblock In {\em Proc.~of the Advances in Neural Information Processing
  Systems (NIPS)}, 1990.

\bibitem{ozkucur2009robocup}
N.E. {\"O}zkucur, B.~Kurt, and H.L. Ak{\i}n.
\newblock {A collaborative multi-robot localization method without robot
  identification}.
\newblock In {\em RoboCup 2008: Robot Soccer World Cup XII 12}, 2009.

\bibitem{prorok2012iros}
A.~Prorok, A.~Bahr, and A.~Martinoli.
\newblock {Low-cost collaborative localization for large-scale multi-robot
  systems}.
\newblock In {\em Proc.~of the IEEE Intl.~Conf.~on Robotics \& Automation
  (ICRA)}, 2012.

\bibitem{prorok2011iros}
A.~Prorok and A.~Martinoli.
\newblock {A reciprocal sampling algorithm for lightweight distributed
  multi-robot localization}.
\newblock In {\em Proc.~of the IEEE/RSJ Intl.~Conf.~on Intelligent Robots and
  Systems (IROS)}, 2011.

\bibitem{ram2011kddm}
P.~Ram and A.G. Gray.
\newblock {Density estimation trees}.
\newblock In {\em Proceedings of the 17th ACM SIGKDD international conference
  on Knowledge discovery and data mining}, 2011.

\bibitem{rohmer2013iros}
E.~{Rohmer}, S.P.N. {Singh}, and M.~{Freese}.
\newblock V-rep: A versatile and scalable robot simulation framework.
\newblock In {\em Proc.~of the IEEE/RSJ Intl.~Conf.~on Intelligent Robots and
  Systems (IROS)}, 2013.

\bibitem{roumeliotis2000tro}
S.I. Roumeliotis and G.A. Bekey.
\newblock {Distributed multi-robot localization}.
\newblock {\em IEEE Trans.~on Robotics (TRO)}, pages 179--188, 2000.

\bibitem{shetty2022iclr}
A.~Shetty, R.~Dwivedi, and L.~Mackey.
\newblock {Distribution Compression in Near-linear Time}.
\newblock In {\em Proc.~of the Int.~Conf.~on Learning Representations (ICLR)},
  2022.

\bibitem{sodano2023icra}
M.~Sodano, F.~Magistri, T.~Guadagnino, J.~Behley, and C.~Stachniss.
\newblock {Robust double-encoder network for rgb-d panoptic segmentation}.
\newblock In {\em Proc.~of the IEEE Intl.~Conf.~on Robotics \& Automation
  (ICRA)}, 2023.

\bibitem{thrun2005probrobbook}
S.~Thrun, W.~Burgard, and D.~Fox.
\newblock {\em {Probabilistic Robotics}}.
\newblock MIT Press, 2005.

\bibitem{wu2009icrb}
D.~Wu and H.~Su.
\newblock {An improved probabilistic approach for collaborative multi-robot
  localization}.
\newblock In {\em 2008 IEEE International Conference on Robotics and
  Biomimetics}, 2009.

\bibitem{zimmerman2023ral}
N.~Zimmerman, T.~Guadagnino, X.~Chen, J.~Behley, and C.~Stachniss.
\newblock {Long-Term Localization Using Semantic Cues in Floor Plan Maps}.
\newblock {\em IEEE Robotics and Automation Letters (RA-L)}, 8(1):176--183,
  2023.

\bibitem{zimmerman2023iros}
N.~Zimmerman, M.~Sodano, E.~Marks, J.~Behley, and C.~Stachniss.
\newblock {Constructing Metric-Semantic Maps Using Floor Plan Priors for
  Long-Term Indoor Localization}.
\newblock In {\em Proc.~of the IEEE/RSJ Intl.~Conf.~on Intelligent Robots and
  Systems (IROS)}, 2023.

\bibitem{zimmerman2022iros}
N.~Zimmerman, L.~Wiesmann, T.~Guadagnino, T.~Läbe, J.~Behley, and
  C.~Stachniss.
\newblock {Robust Onboard Localization in Changing Environments Exploiting Text
  Spotting}.
\newblock {\em Proc.~of the IEEE/RSJ Intl.~Conf.~on Intelligent Robots and
  Systems (IROS)}, 2022.

\end{thebibliography}

\end{document}